\documentclass{article}
\usepackage{pdflscape}

\usepackage{arxiv}

\usepackage[utf8]{inputenc} 
\usepackage[T1]{fontenc}    
\usepackage{url}           
\usepackage{booktabs}       
\usepackage{amsfonts}       
\usepackage{nicefrac}       
\usepackage{microtype}      
\usepackage{lipsum}
\usepackage{multirow}
\usepackage{xcolor}
\usepackage{graphicx}
\usepackage{orcidlink}
\usepackage{comment}
\usepackage{tcolorbox}
\usepackage{array}
\newcolumntype{R}[1]{>{\raggedleft\arraybackslash}p{#1}}
\newcolumntype{L}[1]{>{\raggedright\arraybackslash}p{#1}}

\usepackage{natbib}
\usepackage{longtable}
\usepackage{caption}
\usepackage{amsmath}
\usepackage{rotating}
\usepackage{cleveref}

\makeatletter
\renewcommand{\footnotesize}{\@setfontsize\footnotesize{8pt}{10pt}}
\makeatother

\makeatletter
\renewcommand{\scriptsize}{\@setfontsize\scriptsize{7pt}{9pt}}
\makeatother

\definecolor{VUB_blauw}{rgb}{0.1529, 0.2667, 0.5529}
\usepackage{hyperref}      
\hypersetup{
    colorlinks,
    citecolor=VUB_blauw,
    filecolor=VUB_blauw,
    linkcolor=VUB_blauw,
    urlcolor=VUB_blauw
}
\graphicspath{ {./images/} }
\newcommand{\customCor}[1]{
  \includegraphics[height=1em]{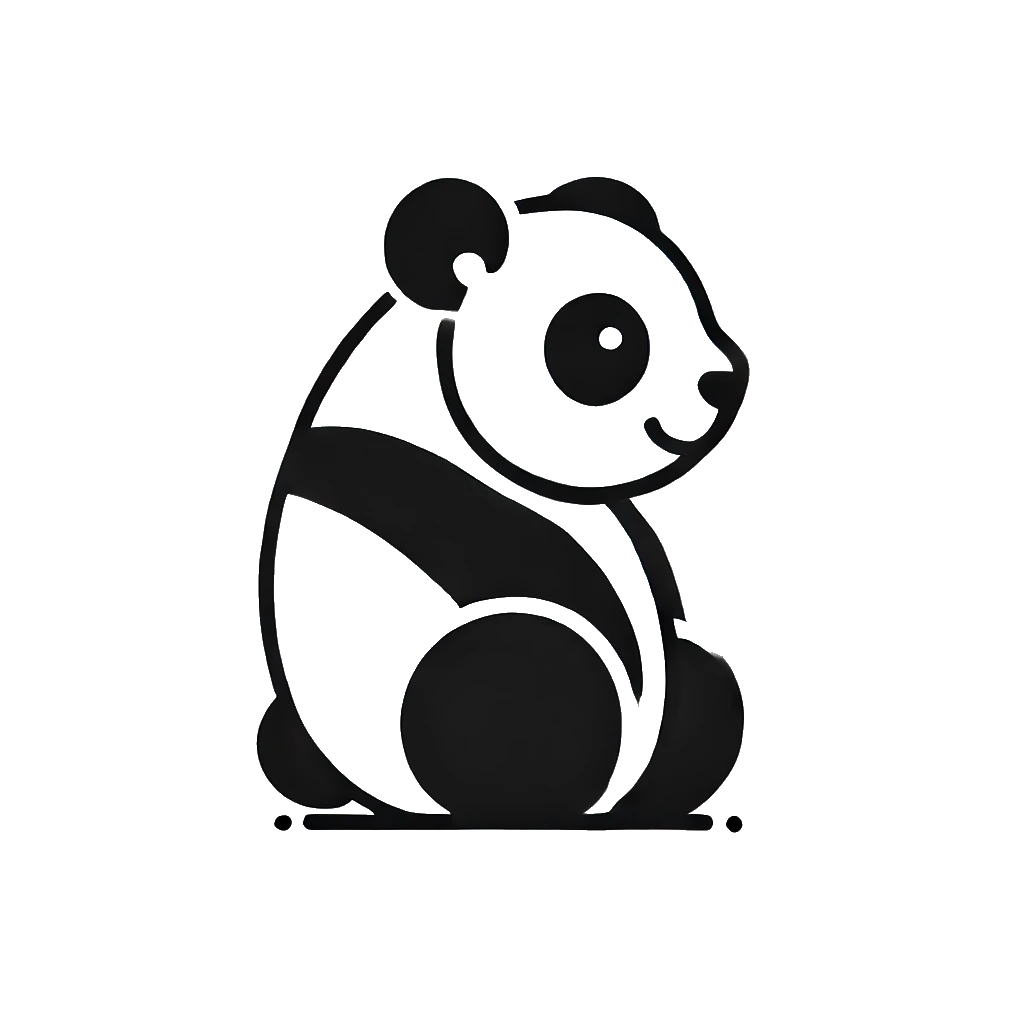} #1%
}

\usetikzlibrary{matrix,calc,shapes.callouts}
\usepackage{tikz}
\tikzset{
  axes/.style    ={line width=0.5pt},          
  gridlines/.style={line width=0.2pt, densely dotted} 
}

\newcommand{\QuadBBox}{
  \path[use as bounding box] (-2,-2) rectangle (2,2);
}

\newcommand{\PlotA}{%
\begin{tikzpicture}
  \QuadBBox
  \draw[gridlines, xstep=.5, ystep=.5] (-1,-1) grid (1,1);
    \draw[axes,<->] (-1.75,0) -- node[above,pos=0.225, anchor=east, yshift=4pt] {\scriptsize \textcolor{white}{p}Performance\textcolor{white}{p}} node[above,pos=0.775, anchor=west, yshift=4pt] {\scriptsize \textcolor{white}{p}Judgement\textcolor{white}{p}} (1.75,0);
    \draw[axes,<->] (0,-1.75) -- node[right,pos=0.02] {\scriptsize \,LLM}  node[right,pos=0.98] {\scriptsize \,Human} (0,1.75);
    \node[
        draw,
        fill=white,
        font = \footnotesize,
        line width = 0.2pt,
        rounded corners=3pt,
        rectangle callout,
        callout absolute pointer={(-1.1,1)},
        inner sep=1.6mm,
        text width=1.8cm,
        align=center,
        line join=miter,        
        miter limit=1,         
        callout pointer width=8pt 
    ] at (-1.6,1.5) {Who and how?};
    \node[anchor=south, font=\bfseries\footnotesize]
  at ([yshift=1mm]current bounding box.north) {\textbf{}Construction\textbf{}};
    \node[font =\scriptsize] at (0.5,0.5) {A};
    \node[font =\scriptsize] at (-0.5,-0.5) {D};
    \node[font =\scriptsize] at (0.5,-0.5) {C,E};
    \node[font =\scriptsize] at (-0.5,0.5) {B};
    \begin{scope}
        \fill[teal, fill opacity=0.07, draw=none] (1,-1) rectangle (0,0);
    \end{scope}
\end{tikzpicture}%
}

\newcommand{\PlotB}{%
\begin{tikzpicture}
  \QuadBBox
  \draw[gridlines, xstep=.5, ystep=.5] (-1,-1) grid (1,1);
    \draw[axes,<->] (-1.75,0) -- node[above,pos=0.225, anchor=east, yshift=4pt] {\scriptsize \textcolor{white}{p}Discrete\textcolor{white}{p}} node[above,pos=0.775, anchor=west, yshift=4pt] {\scriptsize \textcolor{white}{p}Continuous\textcolor{white}{p}} (1.75,0);
    \draw[axes,<->] (0,-1.75) -- node[right,pos=0.02] {\scriptsize \,Static}  node[right,pos=0.98] {\scriptsize \,Dynamic} (0,1.75);
    \node[
        draw,
        fill=white,
        font = \footnotesize,
        line width = 0.2pt,
        rounded corners=3pt,
        rectangle callout,
        callout absolute pointer={(-1.1,1)},
        inner sep=1.6mm,
        text width=1.8cm,
        align=center,
        line join=miter,        
        miter limit=1,         
        callout pointer width=8pt
    ] at (-1.6,1.5) {What form?};
    \node[anchor=south, font=\bfseries\footnotesize]
  at ([yshift=1mm]current bounding box.north) {\textcolor{white}{p}Scale\textcolor{white}{p}};
    \node[font =\scriptsize] at (0.5,0.5) {E};
    \node[font =\scriptsize] at (-0.5,-0.5) {A,B,C,D};
    \begin{scope}
        \fill[teal, fill opacity=0.07, draw=none] (1,1) rectangle (0,0);
    \end{scope}
\end{tikzpicture}%
}

\newcommand{\PlotC}{%
\begin{tikzpicture}
  \QuadBBox
\draw[gridlines, xstep=.5, ystep=.5] (-1,-1) grid (1,1);
    \draw[axes,<->] (-1.75,0) -- node[above,pos=0.225, anchor=east, yshift=4pt] {\scriptsize \textcolor{white}{p}Model-agnostic\textcolor{white}{p}} node[above,pos=0.775, anchor=west, yshift=4pt] {\scriptsize \textcolor{white}{p}Model-specific\textcolor{white}{p}} (1.75,0);
    \draw[axes,<->] (0,-1.75) -- node[right,pos=0.02] {\scriptsize \,No ground truth}  node[right,pos=0.98] {\scriptsize \,Ground truth} (0,1.75);
    \node[
        draw,
        fill=white,
        font = \footnotesize,
        line width = 0.2pt,
        rounded corners=3pt,
        rectangle callout,
        callout absolute pointer={(-1.1,1)},
        inner sep=1.6mm,
        text width=1.8cm,
        align=center,
        line join=miter,        
        miter limit=1,         
        callout pointer width=8pt 
    ] at (-1.6,1.5) {Need anchor?};
    \node[anchor=south, font=\bfseries\footnotesize]
  at ([yshift=1mm]current bounding box.north) {Dependence};
    \node[font =\scriptsize] at (0.5,0.5) {C,D};
    \node[font =\scriptsize] at (-0.5,-0.5) {E};
    \node[font =\scriptsize] at (0.5,-0.5) {};
    \node[font =\scriptsize] at (-0.5,+0.5) {A,B};
    \begin{scope}
        \fill[teal, fill opacity=0.07, draw=none] (-1,-1) rectangle (0,0);
    \end{scope}
\end{tikzpicture}%
}

\AddToHook{shipout/background}{%
  \ifnum\value{page}=1 
    \pagenumbering{Roman} 
    \setcounter{page}{1} 
  \fi
  \ifnum\value{page}=2 
  \pagenumbering{arabic} 
  \setcounter{page}{2} 
  \fi
}

\title{Estimating problem difficulty without ground truth using Large Language Model comparisons}
\runningtitle{}

\author{
 Marthe Ballon\textsuperscript{1, 2, \customCor{ }} \\
    \orcidlinkc{0009-0000-4586-234X} \And
      Andres Algaba\textsuperscript{1,2} \\ 
 \orcidlinkc{0000-0002-0532-3066}\\
 \And
Brecht Verbeken\textsuperscript{1,2} \\
\orcidlinkc{0000-0002-7506-3298}
  \And
Vincent Ginis\textsuperscript{1,2,3} \\ 
 \orcidlinkc{0000-0003-0063-9608} 
 \AND 
  \textsuperscript{1}Data Analytics Lab, Vrije Universiteit Brussel, Pleinlaan 5, 1050 Brussel, Belgium \\
  \textsuperscript{2}imec-SMIT, Vrije Universiteit Brussel, Pleinlaan 9, 1050 Brussels, Belgium\\
  \textsuperscript{3}School of Engineering and Applied Sciences, Harvard University, Cambridge, Massachusetts 02138, USA
}
    
\begin{document}
\maketitle
\renewcommand{\thefootnote}{}
\footnotetext{\includegraphics[height=1em]{panda2.png} Corresponding author: \href{mailto:marthe.ballon@vub.be}{marthe.ballon@vub.be} \\}
\renewcommand{\thefootnote}{\arabic{footnote}}
\thispagestyle{plain}

\begin{abstract}
Recent advances in the finetuning of large language models (LLMs) have significantly improved their performance on established benchmarks, emphasizing the need for increasingly difficult, synthetic data. A key step in this data generation pipeline is a method for estimating problem difficulty. Current approaches, such as human calibration or performance-based scoring, fail to generalize to out-of-distribution problems, i.e. problems currently unsolvable by humans and LLMs, because they are not scalable, time-consuming, and ground truth dependent. Therefore, we propose a new method for estimating problem difficulty, LLM compare, that addresses these limitations. An LLM performs pairwise difficulty comparisons, and then Bradley-Terry scores are computed based on the outcomes. To validate our method, we first propose a conceptual framework that positions existing approaches on three orthogonal planes—construction, scale and dependence—identifying which quadrants a measure needs to occupy to score out-of-distribution problems. LLM compare naturally occupies all desirable quadrants as the first measure that is continuous and dynamic, model-agnostic and independent of ground truth information. As a second validation, we show that LLM compare demonstrates strong alignment with human annotations: Pearson $r\geq 0.80$ for $n=1876$. Thirdly, we show that LLM compare is robust to hallucinations, with less than $6\%$ degradation in Pearson correlation for $10\%$~noise injection. Our work represents a significant step towards replacing time-consuming human annotations and synthetic data generation, and will be an important driver for curriculum design, model evaluation, and AI-assisted research ideation.
\end{abstract}

\keywords{problem difficulty \and ranking \and mathematics \and comprehensive reading  \and large language models \and synthetic data generation}

\section{Introduction}
\label{sec:Introduction}
Scaling the finetuning of large language models (LLMs) marks a step change in their capabilities, as it consistently raises the overall performance on established benchmarks \cite{ballon2025relationship,guo2025deepseek,jaech2024openai,muennighoff2025s1simpletesttimescaling,snell2024scaling,wei2022chain}. Current state-of-the-art models have already saturated long-standing benchmarks like MATH \cite{hendrycks2021measuring} and GSM8K \cite{mirzadeh2024gsm}. Newer benchmarks like Omni-Math \cite{gao2024omni} and GPQA Diamond \cite{rein2023gpqa} are more challenging, yet are expected to reach saturation soon and already raise concerns about data leakage. Recently introduced datasets such as FrontierMath \cite{glazer2024frontiermath} and Humanity’s Last Exam \cite{phan2025humanity} attempt to outpace model capabilities through expert-authored questions, but unfortunately their manual collection process is not scalable. The limited number of high-quality and sufficiently difficult question-response pairs thus necessitates the creation of synthetically LLM-generated data~\cite{setlur2024rl,sun2025self}. Here, the ultimate goal is to create synthetic out-of-distribution data, i.e. problems that are currently unsolvable by both humans and LLMs. In this setting, a key challenge is finding a method for estimating problem difficulty.

The concept of problem difficulty has been approached and measured in various ways. Traditionally, problem difficulty was measured in a human-centric way. In educational assessment, Classical Test Theory (CTT) and Item Response Theory (IRT) define difficulty as a statistical property of test items \cite{hambleton1993comparison,benedetto2023survey}, typically represented by student performance scores. Bloom's Taxonomy and NASA's Task Load Index provide difficulty assessment frameworks to categorize tasks according to their cognitive workload level~\cite{adams2015bloom,hart2006nasa,hart1988development}. Other approaches include manual calibration through expert judgment \cite{bramley2016maintaining} or pre-testing~\cite{lane2015test}, and comparative judgment techniques (pairwise item comparisons), which often yield more reliable difficulty rankings than independent item ratings~\cite{attali2014estimating}. However, as these methods fundamentally depend on human responses, they do not generalize to newly generated, out-of-distribution problems.

In mathematics, economics, and computer science, researchers have proposed formal measures to quantify how challenging a problem is to solve~\cite{calude2006new,cormen2022introduction,page1996two}, yet these measures often have a very narrow scope (e.g. the number of counter-examples necessary to disprove a theorem) or are incomputable.

 The ability of LLMs to assess the inherent difficulty of problems has received little attention, as they often display poor calibration, reporting high confidence while achieving relatively low accuracy \cite{phan2025humanity,wei2024measuring}. Nevertheless, LLM verifiers are already incorporated into the reinforcement learning pipeline for judging final responses or solution processes \cite{guo2025deepseek,setlur2024rewarding,zhang2024generative}, and there is also evidence that they demonstrate promising self-evaluation skills in open-ended sampling tasks \cite{kadavath2022language}. Furthermore, LLMs have been successfully used to assess problem difficulty in various scenarios, e.g. aggregated LLM performance scores \cite{deng2024easy2hard}, supervised learning from question text \cite{benedetto2023survey}, and in-context difficulty learning \cite{gao2024omni}. Unfortunately, these LLM-based approaches depend on specific models, context information, and external calibration, limiting their applicability.

\paragraph{Overview}  In this paper, we present LLM compare, a new method for estimating problem difficulty that does not rely on ground truth data, such as performance scores or reference answers. In this method, an LLM is repeatedly asked to compare two problems and then difficulty scores are computed based on the outcomes, using the Bradley–Terry (BT) model \cite{bradley1952rank}. LLMs have already shown potential in pairwise preference aggregation \citep{chiang2024arena,daynauth2024ranking,wang2024make}, and BT-based scoring has been proven reliable in aggregating LLM judgments \cite{gottweis2025towards,shi2025models,zhang2025replication}. To validate our method, we first propose a new taxonomy that positions existing difficulty measures along three orthogonal planes, related to their construction, scale, and dependence. While prior surveys presented taxonomies of pipelines to supervised, text-based difficulty prediction \cite{alkhuzaey2024text,benedetto2023survey}, our framework classifies the inherent properties of difficulty measures. 
Secondly, we compare our method to four types of existing difficulty measures, human labels, human performance, LLM labels and LLM performance, across three datasets: JEE Advanced Maths 2024, The Cambridge MCQ Reading Dataset and the subset of algebra questions from Omni-Math. We execute the LLM comparisons with both OpenAI o3 and Gemini 2.5 pro to assess model dependence.

\paragraph{Conclusion} 
LLM compare is a fine-grained, reliable and broadly applicable alternative for existing measures of difficulty, resolving their key limitations. We show that LLM compare naturally occupies two empty quadrants in the space of difficulty measures: it is the first measure to be both continuous and dynamic, as well as model-agnostic and independent of ground truth information. This makes it the only measure currently suitable for scoring synthetic out-of-distribution problems. Furthermore, we demonstrate that our method correlates positively with all existing types of difficulty measures, across datasets and models. We also show that LLM compare is the most robust to hallucinations of all LLM-based measures, marking an important step towards solving the problem of time-consuming human annotations and context-dependent LLM-based approaches.

LLM compare provides a key step towards (superhuman) synthetic data generation with LLMs. Our findings highlight the potential of LLMs as difficulty assessors and have implications for curriculum and game design, model evaluation, adaptive testing and AI-assisted research ideation.

\section{LLM compare: a new method for estimating problem difficulty}
\label{sec:LLMcompare}

LLM compare consists of two parts. First, an LLM plays a fixed number of matches per problem in a dataset we want to score. This entails simply letting it decide which out of two problems is more difficult with a basic prompt (see \Cref{subsec:Prompts}). Then, based on the number of wins per question, the difficulty score of each question is estimated by its strength under the Bradley-Terry (BT) model (see \Cref{subsec:BT}).

\subsection{Comparing difficulty with LLMs}
\label{subsec:Compare}

In the first part of LLM compare, we employ OpenAI o3 and Gemini 2.5 Pro to compare the problem pairs and select three datasets to demonstrate the method: JEE Advanced Maths 2024 (JEE), The Cambridge MCQ Reading Dataset (CMCQRD) and the subset of algebra questions from Omni-Math (Omni-Math). Consult \Cref{sec:Datasets} for dataset details. For JEE ($n=34$), we only need to play $561$ matches such that each problem is compared against each other problem. For CMCQRD ($n=787$) and Omni-Math ($n=1876$) this number is much higher: $309,291$ and $1,758,750$, respectively. Therefore, we use the smaller model gpt-oss-120b to test what number of matches per question is required for the BT scores to converge (see \Cref{fig:FigureA4}). In terms of time and expense, we set $\#M=200$ as the maximum number of matches that can be played. For CMCQRD, 66 matches per questions lead to difficulty scores with a Kendall correlation larger than $0.90$ (and other coefficients larger than $0.95$) to $\#M=200$. For Omni-Math, this is $36$ matches per question. This results in $25,971$ and $33,768$ matches played in total by the state-of-the-art models, respectively.

\subsection{Computing Bradley-Terry scores}
\label{subsec:BT}

The second part of LLM compare involves computing Bradley-Terry \cite{bradley1952rank} scores from the pairwise match data. Suppose we have a dataset of $n$ problems, where $\beta_i\in \mathbb{R}$ is the strength of problem $i$. The BT model assumes that the outcome of the comparison between two problems, $i$ and $j$, is independent of the rest of the comparisons. The probability that problem $i$ beats $j$ is then modeled as
$$ p_{ij} =\frac{e^{\beta_i}}{e^{\beta_i} + e^{\beta_j}}.$$
The total number of matches and the pairwise outcomes are known (see \Cref{subsec:Compare}). Hence, under the BT model, we can estimate the strengths $(\beta_1, \beta_2, \dots, \beta_n)$ using iterative Luce Spectral Ranking (ISLR) \cite{NIPS2015_2a38a4a9}. ISLR is a simple, computationally efficient spectral algorithm that converges to the maximum-likelihood estimator for Plackett-Luce/Bradley-Terry models (It comes down to finding the stationary distribution of a Markov chain parameterized by the observed choices). ISLR ($\alpha=0.01$) provides stable, scalable BT strengths from large, sparse LLM pairwise judgments. These strengths act as the final difficulty scores for the $n$ problems. The higher the BT score is, the higher the inferred difficulty. Note that the differences in difficulty between problems are interpretable as the BT scores are linear in the log-odds of success, i.e. if $\beta_i -\beta_j = 1$ and $\beta_k-\beta_l=1$ then $p_{ij}=p_{kl}$.
\section{Results}
\label{sec:Results}

This section illustrates the advantages of using LLM compare for estimating problem difficulty. First, we explain conceptually how LLM compare relates to existing difficulty measures, and why it is currently the only measure suitable to score out-of-distribution problems. Secondly, we empirically validate our method by demonstrating that it correlates well with human annotations and  is robust to LLM hallucinations. All the results in this section remain valid when we use another LLM to compute the LLM-based measures, LLM labels and LLM performance (see \Cref{fig:FigureA6}).

\subsection{LLM compare occupies all desirable quadrants in the space of difficulty measures}
\label{subsec:Conceptual}
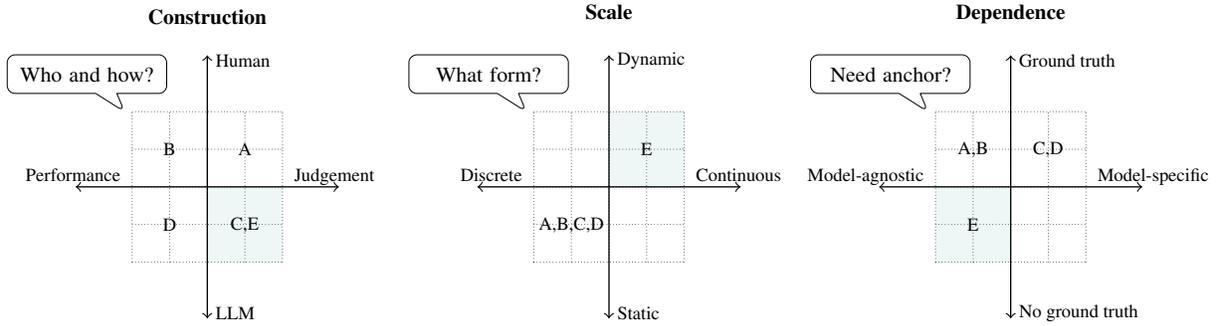
\begin{figure}[h]
\centering
\begin{tikzpicture}[every node/.style={inner sep=0pt,outer sep=0}]
  \matrix[matrix of nodes,
        column sep=0.5in,
          nodes={anchor=center, inner sep=1pt, outer sep=1pt}] (M) {
    \PlotA & \PlotB & \PlotC \\
  };
\end{tikzpicture}
\caption{Conceptually, we can position different measures of problem difficulty along three orthogonal planes. Each plane relates to a different aspect of the measure: how and by whom it is constructed, what form its output takes, and whether it depends on any additional data or models. Every axis is a spectrum on which each difficulty measure can be situated. The following measures are included: A, human labels. B, human performance. C, LLM labels. D, LLM performance. E, LLM compare (our method): We compute Bradley-Terry scores based on matches played by an LLM. The green background hue indicates which quadrants a measure needs to occupy to score superhuman problems.}
\label{fig:Figure1}
\end{figure}

Currently, problem difficulty is measured in various incompatible ways (see \Cref{sec:Introduction}). Therefore, we created a conceptual framework that disentangles the space of difficulty measures (see \Cref{fig:Figure1}). We argue that each difficulty measure can be described based on three separable concepts: \emph{construction} (By whom and how is difficulty evaluated?); \emph{scale} (What form do the output scores take?); and \emph{dependence} (Does the measure need external anchoring?). In this section, we demonstrate how our conceptual framework identifies the exact quadrants that a measure must occupy to score unsolvable problems, and we show that LLM compare naturally occupies these quadrants.

\subsubsection*{The space of difficulty measures}
The space of difficulty measures consists of three orthogonal planes, corresponding to the construction, scale and dependence of the measure (see \Cref{fig:Figure1}). Each axis represents a dimension of the concept, and is regarded as a real axis on which measures can be positioned. 

The first plane, encoding by whom an how a measure is constructed, has dimensions human vs. LLM and performance vs. judgement. For example, Common European Framework of Reference for Languages (CEFR) proficiency levels are human- and judgement based, and formal measures like counting the number of counter examples \cite{calude2006new} are performance- and LLM-based (here, we argue that algorithmic-based measures are somewhat between the LLM and human dimension, illustrating that every axis is a spectrum).

The second plane shows what form the output of the measure takes. It has dimensions dynamic vs. static (with dynamic, we mean the flexibility of the measure to introducing a new problem) and discrete vs. continuous. For example, LLM performance scores are a discrete and static measure for difficulty.

The third plane is related to the dependence of a measure to external anchoring, encoding if the measure depends on a specific model and if it needs context information to provide non-trivial difficulty scores. The corresponding dimensions are ground truth vs. no ground truth and model-agnostic vs. model-specific. Student's pass rates on an exam are model-agnostic but they require ground truth information (solutions to the exam questions).

\subsubsection*{Unexplored quadrants}
All existing methods of determining problem difficulty can be classified into one of four types: human labels (A), human performance (B), LLM labels (C), and LLM performance (D). These types represent the fundamental construct on which the difficulty measure is based. For example, difficulty assignments like IRT or CTT (see \Cref{sec:Introduction}), all aggregate human performance scores in some way, and are therefore fundamentally based on human performance (B). 

Positioning the existing difficulty measures (A–D) in our framework reveals unexplored quadrants in the space of difficulty measures (see \Cref{fig:Figure1}). While all existing measures share the properties of being static, discrete and dependent on ground truth, a universal measure that accommodates for superhuman questions should be dynamic and continuous to be more robust, and not dependent on ground truth as it is unavailable in this setting. Furthermore, such a measure should be LLM- and judgement-based, and preferably model-agnostic. The desirable quadrants are indicated with a green background in \Cref{fig:Figure1}. 

\subsubsection*{Positioning LLM compare in the space of difficulty measures}

LLM compare naturally occupies two empty quadrants in the space of difficulty measures (see \Cref{fig:Figure1}). It is the first measure that is dynamic (with each new question that is scored, the BT scores recalibrate) and whose output is on a continuous scale. It does not depend on a specific model (see \Cref{subsec:model-ag}) or any ground truth information (we use a basic prompt, see \Cref{subsec:Prompts}). By definition, LLM compare is also LLM- and judgement-based, which makes LLM compare the measure uniquely positioned for scoring new generations of benchmarks.

\subsection{LLM compare is aligned with human annotations and robust to hallucinations}
This section shows that LLM compare is a fine-grained measure, that is reliable and generalizable across models and domains. First, we show that LLM compare correlates positively with all existing types of difficulty measures, and most importantly, that it correlates well with human labels (type A). Secondly, we show that our method is robust to LLM hallucinations and independent of the chosen underlying language model. Finally, we confirm that LLM compare is able to identify the presence of difficulty tiers in a benchmark.
\begin{landscape}
    \begin{figure}
        \centering
        \includegraphics{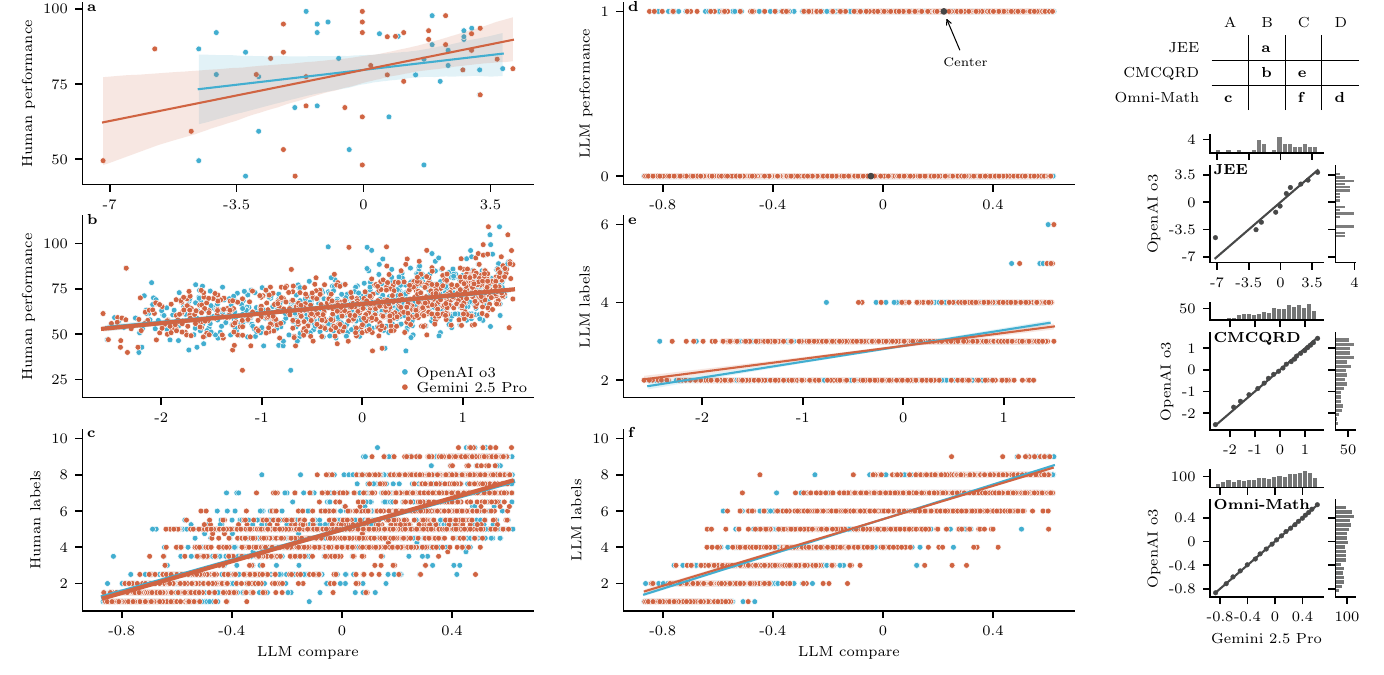}
        \caption{Across datasets, problems deemed more difficult by LLM compare are also deemed more difficult by human labels, human performance, LLM labels and LLM performance. Furthermore, LLM compare is model-agnostic for all datasets. The scatter plots with regression lines (95\% confidence intervals are invisible in some figures) compare the LLM compare difficulty scores for OpenAI o3 and Gemini 2.5 Pro with the four existing methods (see \Cref{fig:Figure1}) across JEE, CMCQRD and Omni-Math. We included all available methods for each dataset (see \Cref{sec:Datasets}). The QQ plots show the agreement between OpenAI o3 and Gemini 2.5 Pro for all three datasets. Note that the spread of the LLM compare difficulty scores is influenced by the total amount of matches played.}
        \label{fig:Figure2}
    \end{figure}
\end{landscape}

\subsubsection*{Correlation with existing difficulty measures}
\label{subsec:correlation}

Across all datasets, problems that LLM compare considers more difficult are also considered more difficult by existing difficulty measures. Panels a$-$f of \Cref{fig:Figure2} show a positive correlation between LLM compare on the one hand and human labels, human performance, LLM labels and LLM performance on the other hand (the table in the top right-hand corner indicates which dataset was used for which panel). This trend is confirmed by the Pearson, Spearman and Kendall correlation coefficients, summarized in \Cref{tab:Table1}. Note that the Kendall $\tau$ lies naturally lower for all difficulty measures because it punishes discordant pairs and ties explicitly.

The number of ties does not dominate the strength of the correlation. The presence of ties in the score of the discrete measures ensures that the correlation with LLM compare is always less than $1$, even for a perfect monotonic relationship. However, when you combine the scatter plots in \Cref{fig:Figure2} with the data in \Cref{tab:Table1}, it is clear that a lower correlation is primarily due to a misalignment in the difficulty scores rather than the degree of discreteness of the measure.  

\subsubsection*{Robustness to hallucinations}
\label{subsec:robust}

\begin{figure}
    \centering
    \includegraphics{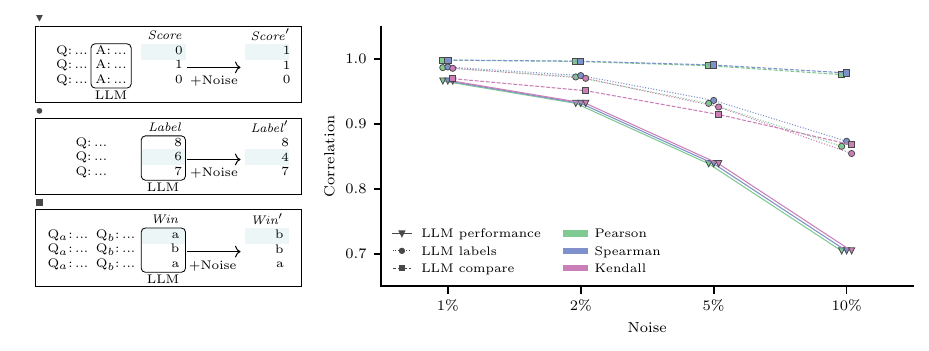}
    \caption{Of the three LLM-based difficulty measures, LLM compare is the most robust to hallucinations. This is primarily because it is a more fine-grained measure. (Left) We demonstrate conceptually how we introduce noise (i.e. LLM hallucinations) into the LLM performance, LLM labels and LLM compare scores. The same percentage of noise was added at the LLM stage for the three methods, whereby a wrong answer was made correct or vice versa, a wrong label was assigned to a question, the wrong question was picked in the difficulty comparison, respectively. (Right) We plot the mean Pearson, Spearman and Kendall correlations between the original difficulty score and the noisy data (1000 samples) for noise levels $1\%$, $2\%$, $5\%$ and $10\%$. The $95\%$ confidence intervals are omitted from the figure as they were negligibly small. The level of hallucination has almost no influence on the Pearson and Spearman coefficient for the LLM compare method, while the other methods degrade quickly. As the LLM compare scores are on a continuous scale and do not have ex aequo's as a result, the Kendall tau deviates from this trend.}
    \label{fig:Figure3}
\end{figure}

Of all the LLM-based difficulty measures, LLM compare is the most robust to hallucinations. We introduced a percentage of hallucinations to LLM performance, LLM labels and LLM compare in the following way: a wrong answer was made correct and vice versa; a wrong label was assigned to a problem; and the wrong problem was selected in the difficulty comparison (left panel of \Cref{fig:Figure3}). The right-hand panel of \Cref{fig:Figure3} displays the correlation between the original difficulty score assigned by the three LLM-based measures and the noisy score, for noise levels $1\%$, $2\%$, $5\%$ and $10\%$. We again include Pearson, Spearman and Kendall coefficients. According to all three correlation metrics, the LLM compare difficulty scores degrade the least compared to the original sequence. The Kendall coefficient is smaller for LLM compare than for LLM labels at low noise levels. This is because LLM compare is a continuous measure, so each hallucination creates a discordant pair, whereas rank could still be maintained in LLM labels (the same holds for LLM performance). Note that when the sequence is binary, the three correlation coefficients coincide.

\subsubsection*{Independence of underlying language model}
\label{subsec:model-ag}

LLM compare is model-agnostic. We used two state-of-the-art LLMs--OpenAI o3 and Gemini 2.5 Pro--to execute the pairwise comparisons in our method (see \Cref{subsec:Compare}). For sufficiently large datasets, the scatter plots in \Cref{fig:Figure2} already demonstrate a high degree of agreement between OpenAI o3 and Gemini 2.5 Pro. Furthermore, the QQ plots on the right-hand side of \Cref{fig:Figure2} show almost perfect alignment of their BT scores, for all three datasets. Note that the spread of BT scores is influenced by the total number of matched played, as shown in the corresponding histograms.

\subsubsection*{Ability to identify difficulty tiers}
\label{subsec:clusters}
\begin{figure}
    \centering
    \includegraphics{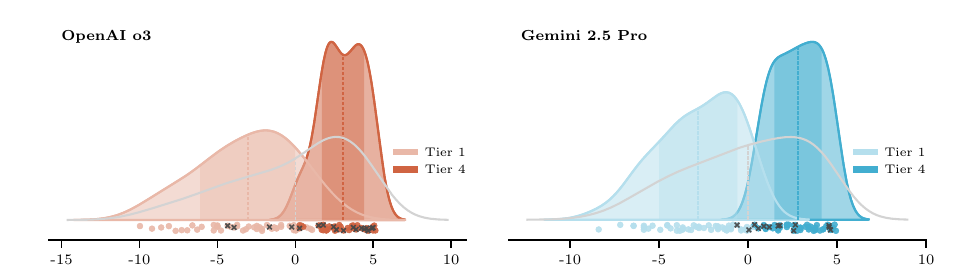}
    \caption{LLM compare is able to identify the presence of difficulty tiers in a benchmark. We apply LLM compare to a synthetic dataset consisting of 50 problems sampled from Omni-Math Tier $1$ (= easiest problems according to human labels, see \Cref{subsec:OmniMath}) and 50 problems sampled from Tier $4$ (= hardest problems). We use again OpenAI o3 and Gemini 2.5 Pro to perform the LLM comparisons. The kernel density estimate (kde) plots for the BT scores of the two tiers show a mean separation of $6.12$ for OpenAI o3 and $5.61$ for Gemini 2.5 Pro. We further quantify the separation between the Tier 1 and 4 BT score distributions in \Cref{subsec:clusters}. The gray kde-plot shows the entire distribution of LLM compare output scores and is theoretically centered around $0$. The crosses indicate which problems where answered incorrectly by the respective models (using Omni-Judge \cite{gao2024omni}), indicating that problems that the models answered incorrectly are not necessarily judged as the most difficult.}
    \label{fig:4}
\end{figure}

BT scores are able to identify the presence of difficulty tiers in a benchmark. The BT scores are linear in the log-odds of success, making LLM compare also a measure for relative difficulty between the problems in a benchmark (see \Cref{subsec:BT}). This is especially relevant when a benchmark consists of groups of equivalent problems or contains a small set of really hard/easy problems. We applied LLM compare to a synthetic dataset consisting of $50$ Omni-Math Tier $1$ (= easiest) questions and $50$ Tier $4$ (= hardest) questions, again using both OpenAI o3 and Gemini 2.5 Pro (see \Cref{fig:4}). The resulting distributions show a clear separation between the two difficulty tiers. We further quantify LLM compare's ability to separate difficulty tiers using Cohen's $d$ (a measure for the effect size of the mean difference) and the Wasserstein distance (a measure reflecting among other things the area between two empirical cdf's). Cohen's $d$ between Tier $1$ and Tier $4$ for OpenAI o3 is $2.56$ indicating that the mean separation is approximately $2.5$ times the pooled standard deviation of the two distributions. For Gemini 2.5 Pro, we obtain $d=3.06$. The Wasserstein distance relative to the joint range of both distributions is equal to $0.40$ and $0.42$ respectively, showing again a substantial degree of separation between Tier 1 and Tier 4 BT scores.

\section{Discussion}

This work presents LLM compare, a new method for estimating problem difficulty without ground truth. Our method consists of two parts: First, an LLM performs pairwise difficulty comparisons (\Cref{subsec:Compare}). Then Bradley–Terry strengths are fitted to the outcomes of these comparisons (\Cref{subsec:BT}). We validate LLM compare by providing a conceptual taxonomy of difficulty measures, and empirical evidence of its generality and reliability across models and domains.

Our taxonomy (\Cref{subsec:Conceptual}) reveals that LLM compare is uniquely positioned to score out-of-distribution problems, i.e. problems that are currently unsolvable by both humans and LLMs. We disentangle the space of difficulty measures along three orthogonal planes, related to the construction, scale, and dependence of the measure. Each axis represents a dimension (e.g., human vs. LLM) on which difficulty measures can be positioned. By positioning the four existing types of difficulty measures—human labels, human performance, LLM labels, LLM performance—we can identify which quadrants a measure needs to occupy to score out-of-distribution problems. As the first measure that is dynamic and continuous, independent of ground-truth information (\Cref{subsec:Prompts}) and model-agnostic (\Cref{subsec:model-ag}), LLM compare naturally occupies all desirable quadrants.  

LLM compare correlates positively with all types of existing difficulty measures and is the most robust to hallucinations out of all LLM-based measures. We compared LLM compare to six difficulty measures extracted from three datasets: JEE Advanced Maths 2024, The Cambridge MCQ Reading Dataset, and the subset of algebra question from Omni-Math. A visual inspection of the relationship between every difficulty score and LLM compare showed good alignment across datasets and models (\Cref{subsec:correlation}). This was confirmed by positive Pearson, Spearman, and Kendall correlations, most notably a Pearson correlation of over $80\%$ with human labels. To test robustness, we added $1$-$10\%$ hallucinations to each LLM-based measure and found that LLM compare degraded the least relative to the original sequence (\Cref{subsec:robust}).

\paragraph{Practical implications} LLM compare is an important step towards solving the need for time-consuming human annotations as it shows high correlation to human labels and is robust to hallucinations. Furthermore, the BT scores reflect the presence of tiers in benchmark difficulty, e.g. when a dataset contains a group of equivalent problems (\Cref{subsec:clusters}). The relative differences in difficulty between problems are also interpretable (\Cref{subsec:BT}), in contrast to ranking- or binning-based difficulty measures.

\paragraph{Limitations and future work}
Our findings address a key challenge in the pipeline of generating increasingly difficult synthetic data.
However, effectively creating a superhuman dataset, and verifying the insolvability of the problems is not entirely straightforward. The discriminator in our pipeline is LLM compare, which we can apply in an iterative-refinement setting (Is this question difficult enough?) or in a selective setting (Out of these generated questions, which ones are the most difficult?). Note that we can also verify other concepts with our method (e.g. solvability or relevance). Creating an appropriate generator is ongoing work. 
\Cref{fig:Figure2} shows that OpenAI o3 and Gemini 2.5 Pro correlate strongly on the three considered datasets. However, in \Cref{subsec:Compare}, we assume that this also holds for gpt-oss-120b (i.e. that the BT scores converge for the same number of matches per problem as gpt-oss-120b). One final limitation is that the employed datasets span only 2 disciplines (math and reading comprehension) and that state-of-the-art LLMs achieve high accuracy on all three of them. This underscores the significance of our approach again, as it is not limited by available human- or performance-based data.

\newpage
\section{Materials and Methods}

\subsection{Datasets}
\label{sec:Datasets}

To demonstrate how LLM compare works, we compared it to the four existing measures identified in \Cref{subsec:Conceptual}: human labels (A), human performance (B), LLM labels (C) and LLM performance (D). Although LLM compare can be applied to any set of questions, existing benchmarks are usually not accompanied by all four types of difficulty scores. Therefore, we selected three datasets for which at least one type is available: JEE Advanced Maths 2024 (JEE), The Cambridge MCQ Reading Dataset (CMCQRD) and the subset of algebra questions from Omni-Math (Omni-Math). This section elaborates on the content of the datasets and the process of extracting the difficulty scores.

\subsubsection*{JEE Advanced Maths 2024}

The Joint Entrance Examination Advanced is a prestigious annual examination held in India that tests applicants' knowledge of chemistry, physics and mathematics. We select the mathematics questions of the $2024$ examination, resulting in a dataset of $34$ mathematics questions at a competitive high school level (Paper $1$ and $2$ contain $17$ questions each \cite{JIC-JEE-Advanced-2024}).

\textbf{Human performance}\quad The JEE annual report \cite{JIC-JEE-Advanced-2024} contains question-wise statistics on the participants' performance. From the percentages of fully correct answers and partially correct answers, we computed a new score called EffectiveCorrect, where partial correct responses contribute with weight factor $\frac{1}{2}$. This is in line with the official scoring mechanism. The difficulty of each question is then defined as $100$-$\%$EffectiveCorrect. 

As the JEE dataset is a small dataset which is saturated by state-of-the-art models, we have chosen not to include LLM labels or LLM performance.  (By "saturated", we mean a dataset for which state-of-the-art models achieve near-perfect accuracy, or for which model errors signal imperfections in the dataset rather than unsolvability.) Human labels are also unavailable for this dataset.

\subsubsection*{The Cambridge MCQ Reading Dataset}
\label{subsec:CMCQRD}

CMCQRD \cite{mullooly2023cambridge} consists of $120$ reading tasks (written text passages) with multiple $4$-option questions assessing the participants' reading comprehension. This results in a dataset with $787$ individual multiple-choice questions (originally, the dataset contained $792$ entries, but we deleted 5 empty cells). It covers multiple CEFR proficiency levels (B1-C2) and a broad-range of reading comprehension topics. 

\textbf{Human performance}\quad The authors of \cite{mullooly2023cambridge} compute a facility, discrimination and difficulty score for each entry in the dataset. For the difficulty score they use the $1$ parameter Rasch model from Item-Response-Theory (IRT) to estimate the difficulty of a problem based on the performance data. After calibration and scaling (such that the scores are approximately between $0$ and $100$), they obtain a performance-based score that also takes into account the participants skill level \cite[Item Difficulty]{mullooly2023cambridge}.

\textbf{LLM labels}\quad We prompt OpenAI o3 and Gemini 2.5 Pro to assign a difficulty score on a scale of $1$ to $10$ without solving the problem (consult \Cref{subsec:Prompts} for prompt details and \Cref{sec:appendix} for results with Gemini 2.5 Pro). We take the raw output values as difficulty scores of type LLM labels. 

There are human labels available for CMCQRD, but they are used to calibrate the IRT scores \cite{mullooly2023cambridge}, hence we do not include them. This dataset is also saturated, thus LLM performance is non-informative.

\subsubsection*{Omni-Math Algebra}
\label{subsec:OmniMath}

The original Omni-Math dataset \cite{gao2024omni} contains $4,428$ Olympiad-level math problems and answers, along with rigorous domain and difficulty classifications (\Cref{fig:sample-3}). The authors of \cite{ballon2026saturate} cleaned the Omni-Math dataset by manually correcting the problem statements for missing info, missing images, invalid latex code, duplicates and degenerate problems. Furthermore, they labeled problems that ask for a proof or an estimation and the ones that contain an image. We use a subset of this cleaned dataset: the set of algebra problems that do not contain proof questions, estimations, or images. This results in a dataset of $1876$ challenging algebra problems, that do not require multi-modal models, is sufficiently large and spans all difficulty levels of the original dataset. 

\textbf{Human labels}\quad Each problem in Omni-Math is annotated with a difficulty score ranging from $1$ to $10$, with increments of $0.25$ and $0.5$. The difficulty scores were assigned as follows: For competitions included on the AoPS forum, expert difficulty ratings are available; therefore, the authors of \cite{gao2024omni} assign the corresponding difficulty score to the problem directly. For competitions not included in the forum, the authors use the existing competitions and their associated difficulties on this page as a basis for in-context learning, prompting GPT-4o to assign specific difficulty ratings. As the GPT-4o prompt contains detailed difficulty references consistent with those on the AoPS forum, and the difficulty scores are consistent with the human ranking of the competitions (see Section 2.2 of \cite{gao2024omni}), we regard the combined difficulty scoring as human labels.

\textbf{LLM labels}\quad We assign difficulty levels with OpenAI o3 and Gemini 2.5 Pro in the same way as for CMCQRD (consult \Cref{subsec:Prompts} for prompt details and \Cref{sec:appendix} for results with Gemini 2.5 Pro).

\textbf{LLM performance}\quad Omni-Math is the only non-saturated dataset used in this paper. OpenAI o3 achieves an $83.5\%$ score on the algebra question subset (Gemini 2.5 Pro achieves $86.5\%$), using Omni-Judge \cite{gao2024omni} as verifier. We use the scoring system where the problem receives a difficulty score of $1$ if the model answers incorrectly and $0$ if it answers correctly, as an LLM performance-based measure. 

Human performance scores are currently not available for this dataset.

\subsection{Models}
The large language models used in this paper are 
\begin{itemize}
    \item OpenAI o3, (\texttt{o3-2025-04-16}, medium reasoning effort, no token limit, OpenAI Batch API)
    \item Gemini 2.5 Pro (\texttt{gemini-2.5-pro}, $8192$ tokens thinking budget, Gemini Batch API)
    \item gpt-oss-120b (\texttt{openai/gpt-oss-120b}, high reasoning effort, OpenRouter)
    \item Omni-Judge (\texttt{kbsdjames.omni-judge}, $4096$ tokens context length, $300$ tokens output budget, LM Studio)
\end{itemize}

\subsection{Correlation coefficients}
We use the built in Pearson, Spearman and Kendall correlation from scipy.stats version 1.16.2. This results in using Kendall's $\tau_b$ by default, hence accounting for ties explicitly.

\subsection{Distance metrics}
To quantify the separation between two distributions of BT scores, we use Cohen's $d$ and the Wasserstein-1 distance between two one-dimensional distributions. Cohen's $d$ is computed by
$$d = \frac{\overline{x}_1-\overline{x}_2}{\sqrt{\frac{(n_1-1)s_1^2+(n_2-1)s_2^2}{n_1+n_2-2}}},$$
where ${x}_1$ and ${x}_2$ are the sequences of BT scores from Tier $1$ and Tier $4$, respectively. The Wasserstein distance is calculated as follows (built in scipy.stats)
 $$W(\mu_1,\mu_2)=\int_{\mathbb{R}}|F_1(x)-F_2(x)|dx,$$
 where $F_1$ and $F_2$ are the cumulative distribution functions corresponding to measures $\mu_1$ and $\mu_2$. Subsequently, we divide $W$ by the joint range of the distributions to make the distance interpretable (i.e. on a scale of $0-1$).

\subsection{Prompts}
\label{subsec:Prompts}

We used the following prompt to execute the pairwise comparisons for both OpenAI o3 and Gemini 2.5 Pro (see \Cref{subsec:Compare}).  
\begin{tcolorbox}
You will be given two problems. Assess which problem is more difficult. Do not solve them. Your final response will be a single letter between the @< and >@ special tokens (i.e., @<a>@ or @<b>@). Problem a: \{problem$_i$\}$\setminus$n Problem b: \{problem$_j$\}
\end{tcolorbox}
\vspace{5mm}

We used the following prompt to benchmark OpenAI o3 and Gemini 2.5 Pro on the subset of algebra questions that do not contain proofs, estimations or images of Omni-Math \cite{gao2024omni} (see \Cref{subsec:OmniMath}). To correct their answers with Omni-Judge, we use the same few-shot prompt as in \cite{gao2024omni}.

\begin{tcolorbox}
You will be given a question. Answer it by providing the final answer between @< and >@ special tokens (e.g., @<2>@, @<n=3>@, @<Yes>@).
\end{tcolorbox}
\vspace{5mm}

We used the following prompt to create LLM labels for CMCQRD and Omni-Math by both OpenAI o3 and Gemini 2.5 Pro (see \Cref{subsec:CMCQRD,subsec:OmniMath}).
\begin{tcolorbox}
    You will be given a question. Determine the difficulty of the question on a scale from 1 to 10. Do not solve it. Your final response will be a single number between <@ and @> special tokens (e.g., <@2@>, <@8@>).
\end{tcolorbox}

\clearpage
\section*{Acknowledgements}
This research received funding from the Flemish Government (AI Research Program).
Andres Algaba acknowledges support from the Francqui Foundation (Belgium) through a Francqui Start-Up Grant and a fellowship from the Research Foundation Flanders (FWO) under Grant No.1286924N. 
Vincent Ginis acknowledges support from Research Foundation Flanders under Grant No.G032822N and G0K9322N. 

\section*{Author contributions}
Vincent Ginis, Andres Algaba and Marthe Ballon were responsible for the main idea of the study. Brecht Verbeken and Andres Algaba implemented the pairwise comparisons and BT-scoring. Marthe Ballon conducted the rest of the analysis, made the figures and drafted the manuscript. All authors collaboratively revised the manuscript and provided critical feedback.

\section*{Data and code availability}
Data associated with this study are available in a public repository at \url{https://doi.org/10.5281/zenodo.17523641}. \\
The JEE Advanced Maths 2024 dataset is available at \hyperlink{https://jeeadv.ac.in/archive.html} {https://jeeadv.ac.in/archive.html}.\\
The Cambridge MCQ Reading Dataset is available at \hyperlink{https://englishlanguageitutoring.com/datasets/cambridge-multiple-choice-questions-reading-dataset}{https://datasets/cambridge-mcq-reading-dataset}. \\
The original Omni-Math dataset is available at \url{https://huggingface.co/datasets/KbsdJames/Omni-MATH}. \\ 

The code for this publication is publicly available at \url{https://github.com/MartheBallon/estimating-problem-difficulty-without-ground-truth}. We used Python 3.12.0 (\textit{choix 0.4.1}, \textit{datasets 4.0.0}, \textit{google-genai  1.29.0}, \textit{matplotlib  3.10.5}, \textit{networkx  3.5}, \textit{numpy  2.3.2}, \textit{openai  1.99.9}, \textit{pandas  2.3.0}, \textit{scipy 1.16.2} and \textit{seaborn  0.13.2}).
\bibliographystyle{unsrt}

\appendix

\setcounter{figure}{0}
\renewcommand{\thefigure}{A\arabic{figure}}
\setcounter{table}{0}
\renewcommand{\thetable}{A\arabic{table}}

\clearpage
\section{Appendix}
\label{sec:appendix}
\vfil
\begin{table*}[h]
\centering
\begin{tabular}{L{4cm}lcccc}
 & \textbf{LLM compare} \qquad \quad & \textbf{Pearson} & \textbf{Spearman} & \textbf{Kendall}  \\
\hline
\addlinespace
\addlinespace
\textbf{Human performance} &  \\
\addlinespace
\multirow{2}{*}{\text{JEE} ($n=34$)} & OpenAI o3  & 0.24 & 0.16 & 0.10 \\
\addlinespace
 & Gemini 2.5 Pro & 0.44 & 0.32 & 0.21  \\
 \addlinespace
 \addlinespace
\multirow{2}{*}{\text{CMCQRD} ($n=787$)} & OpenAI o3  & 0.48 & 0.51 & 0.35  \\
\addlinespace
 & Gemini 2.5 Pro & 0.53 & 0.56 & 0.38  \\
\addlinespace
\addlinespace
\textbf{Human labels} \\
\addlinespace
\multirow{2}{*}{\text{Omni-Math} ($n=1876$)} & OpenAI o3  & 0.80 & 0.78 & 0.60  \\
\addlinespace
 & Gemini 2.5 Pro &  0.82 & 0.80 & 0.62  \\
\addlinespace
\addlinespace
\textbf{LLM performance} \\
\addlinespace
\multirow{2}{*}{\text{Omni-Math} ($n=1876$)} & OpenAI o3  & 0.24 & 0.25 & 0.21  \\
\addlinespace
 & Gemini 2.5 Pro & 0.22 & 0.23 & 0.18  \\
\addlinespace
\addlinespace
\textbf{LLM labels} \\
\addlinespace
 \multirow{2}{*}{CMCQRD ($n=787$)} & OpenAI o3 & 0.61 & 0.60 & 0.49 \\
 \addlinespace
& Gemini 2.5 Pro & 0.69 & 0.74 & 0.61 \\
\addlinespace
\addlinespace
\multirow{2}{*}{\text{Omni-Math} ($n=1876$)} & OpenAI o3 &  0.91 & 0.90 & 0.77 \\
\addlinespace
 & Gemini 2.5 Pro & 0.90 & 0.89 & 0.76 \\
 \addlinespace
\hline
\end{tabular}
\caption{Summary of the comparison between LLM compare (our method) and four existing difficulty measures--human labels, human performance, LLM labels and LLM performance. Pearson denotes Pearson's correlation coefficient, Spearman denotes Spearman's rank correlation coefficient, Kendall denotes Kendall's rank correlation coefficient.}
\label{tab:Table1}
\end{table*}
\clearpage

\begin{figure*}[t]
    \centering
    \scalebox{1}{
    \begin{tcolorbox}[fonttitle=\bfseries,title= Sample problem JEE]

\textbf{Question Number: }Q. $1.2$

\textbf{FullCorrect: }$13.34$

\textbf{PartialCorrect: }$00.00$\newline

\textbf{Problem: } A student appears for a quiz consisting of only true-false type questions and answers all the questions. The student knows the answers of some questions and guesses the answers for the remaining questions. Whenever the student knows the answer of a question, he gives the correct answer. Assume that the probability of the student giving the correct answer for a question, given that he has guessed it, is $\frac{1}{2}$. Also assume that the probability of the answer for a question being guessed, given that the student's answer is correct, is $\frac{1}{6}$. Then the probability that the student knows the answer of a randomly chosen question is (A) $\frac{1}{12}$; (B) $\frac{1}{7}$; (C) $\frac{5}{7}$; (D) $\frac{5}{12}$.\newline

\textbf{Answer: }C\newline
\end{tcolorbox}}
    \caption{Sample problem of the JEE Advanced Maths 2024.}
    \label{fig:sample-1}
\end{figure*}

\begin{figure*}[ht]
    \centering
    \scalebox{1}{
    \begin{tcolorbox}[fonttitle=\bfseries,title= Sample problem CMCQRD]

\textbf{Question Number: }$0$

\textbf{Difficulty: }$83.03$\\

\textbf{Problem: } --- Passage ---
Some time ago a website highlighted the risks of public check-ins – online announcements of your whereabouts. The site’s point was blunt: you may think you are just telling the world, ‘Hey, I’m at this place’ – but you are also advertising your out-and-about-ness to all kinds of people everywhere – not all of them people you might like to bump into. This appeared to confirm the growing awareness that there might be a downside to all the frantic sharing the web has enabled. The vast new opportunities to publish any and every aspect of our lives to a potentially global audience hold out all sorts of tantalising possibilities: Wealth! Fame! So we plunge into the maelstrom of the internet, tossing confessions, personal photos and stories into the digital vortex. Too late we realise that the water is crowded and treacherous – and we are lost. 

Depressing? Perhaps, but don’t give up. This future has a map, drawn for us years ago by a reckless group of online pioneers. In the early days of the web, they sailed these waters and located all the treacherous shoals. They got fired from their jobs, found and lost friends and navigated celebrity’s temptations and perils – all long before the invention of social networking. These pioneers, the first wave of what we now call bloggers, have already been where the rest of us seem to be going. Before their tales scroll off our collective screen, it’s worth spending a little time with them. After all, those who cannot learn from history are doomed to repost it. 

In January 1994, Justin Hall, a 19-year-old student, began posting to the ‘WWW’, as it was then known, something inhabited mostly by grad students, scientists and a handful of precocious teens like him. The web had been invented at Cern, the international physics lab in Switzerland, so researchers could more easily share their work. Hall saw something else: an opportunity to share his life. Link by link, he built a hypertext edifice of autobiography, a dense thicket of verbal self-exposure leavened with photos and art. In January 1996, on a dare, he began posting a daily blog, and readers flocked to the spectacle of a reckless young man pushing the boundaries of this new medium in every direction at once.

Hall’s ethos was absolute: cross his path and you could appear on his site; no topic was taboo. Certainly, this was the work of an exhibitionist, but there was also a rigour and beauty to his project that only a snob would refuse to call art. One day though, visitors to Hall’s site discovered his home page gone, replaced with a single anguished video titled Dark Night. His story tumbled out; he’d fallen spectacularly in love, but when he started writing about it on his site he was told ‘either the blog goes, or I do’. He’d published his life on the internet and, Hall protested, ‘it makes people not trust me’. The blog went, but the dilemma persists. Sharing online is great. But if you expect your song of yourself to ‘make people want to be with you’, you’ll be disappointed. 

In 2002, Heather Armstrong, a young web worker in Los Angeles, had a blog called Dooce. Occasionally, she wrote about her job at a software company. One day an anonymous colleague sent the address of Armstrong’s blog to every vice president at her company – including some whom she’d mocked – and that was the end of her job. Those who study the peculiar social patterns of the networked world have a term to describe what was at work here. They call it the ‘online distribution effect’: that feeling so many of us have that we can get away with saying things online that we’d never dream of saying in person. But the web isn’t some king of alternative reality where we can let our hair down without worrying about repercussions. Our digital lives are interwoven with our real lives. When we pretend otherwise, we risk making terrible, life-changing mistakes.

Armstrong’s saga had a happy ending. Though she was upset by the experience and stopped blogging for several months afterwards, she ended up getting married and restarting her blog with a focus on her new family. Today she is a star in the burgeoning ranks of ‘mommy bloggers’ and her writing supports her household. Once a poster-child for the wages of web indiscretion, she has become a virtuoso of managed self-revelation. What Armstrong has figured out is something we would all do well to remember: the web may allow us to say anything, but that doesn’t mean we should.\\

--- Question ---
Why does the writer describe a website about public check-ins in the first paragraph?\\
a. to reinforce the concerns already felt by some people

b. to remind readers to beware of false promises

c. to explain that such sites often have a hidden agenda

d. to show that the risks of internet use are sometimes over estimated\newline

\textbf{Answer: }a\newline
\end{tcolorbox}}
    \caption{Sample problem of The Cambridge MCQ Reading Dataset.}
    \label{fig:sample-2}
\end{figure*}

\clearpage

\clearpage

\begin{figure*}[t]
    \centering
    \scalebox{1}{
    \begin{tcolorbox}[fonttitle=\bfseries,title= Sample problem Omni-Math]
\textbf{Domain: }Mathematics $\rightarrow$ Algebra $\rightarrow$ Polynomial Operations

\textbf{Difficulty: }$8.5$

\textbf{Source: }China National Olympiad\newline

\textbf{Problem: }Do there exist positive reals $a_0, a_1,\ldots ,a_{19}$, such that the polynomial $$P(x)=x^{20}+a_{19}x^{19}+\ldots +a_1x+a_0$$ does not have any real roots, yet all polynomials formed from swapping any two coefficients $a_i,a_j$ have at least one real root?
 \newline

\textbf{Answer: }$\text{Yes}$\newline

\textbf{Solution: } To determine whether there exist positive reals \(a_0, a_1, \ldots, a_{19}\) such that the polynomial \(P(x) = x^{20} + a_{19}x^{19} + \ldots + a_1x + a_0\) does not have any real roots, yet all polynomials formed from swapping any two coefficients \(a_i, a_j\) have at least one real root, we proceed as follows:

Consider the polynomial \(P_\sigma(x) = x^{20} + a_{\sigma(19)}x^{19} + a_{\sigma(18)}x^{18} + \cdots + a_{\sigma(0)}\), for all permutations \(\sigma\) of the numbers 0 to 19.

We construct the coefficients \(a_i\) in a specific manner. Let \(a_i = 10000 + i\epsilon\) for \(i = 0, 1, \ldots, 19\) and some small \(\epsilon > 0\). This ensures that \(a_0 < a_1 < \cdots < a_{19}\).

When \(t = 0\), we substitute \(x = -100\). Since \(\frac{|a_{19} \cdot 100^{19}|}{20} > |100^{20}|, |a_{18} \cdot 100^{18}|, |a_{17} \cdot 100^{17}|, \ldots, |a_0|\), we have \(P(-100) < 0\).

As \(t \rightarrow \infty\), \(a_{18} \rightarrow \infty\). When \(a_{18} > -\min_{x < 0} \left( x^2 + a_{19}x + \frac{a_{17}}{x} + \cdots + \frac{a_0}{x^{18}} \right)\), \(P(x) \geq 0\) for all \(x < 0\). This minimum exists because as \(x \rightarrow 0\), \(\frac{a_0}{x^{18}}\) dominates and the sum tends to positive infinity, so it is positive for some \(x > x_0\). Meanwhile, as \(x \rightarrow -\infty\), \(x^2\) dominates, and the sum is positive for some \(x < x_1\). The middle interval is closed and bounded, so it achieves its minimum which is finite.

Meanwhile, \(P(x) > 0\) for all \(x \geq 0\).

Fix \(t\) as the minimum value such that \(P(x) \geq 0\) for all \(x\). By continuity, there is a root \(y\) of \(P(x)\), which is clearly negative. If \(-1 \leq y < 0\), then \(a_{19}y^{19} + a_{18}y^{18} > a_{18}(y^{18} + y^{19}) \geq 0\). Grouping the rest similarly in pairs, and using \(y^{20} > 0\), \(P(y) > 0\), a contradiction.

Hence \(y < -1\), and \(y^{19} < y^{17} < \cdots < y^1 < y^0 < y^2 < \cdots < y^{18}\). Since \(a_{19} < a_{17} < \cdots < a_1 < a_0 < a_2 < \cdots < a_{18}\), by the rearrangement inequality, \(0 = P(y) > P_\sigma(y)\) for \(\sigma \neq \text{Id}\).

Adding a small \(\delta\) to \(t\), \(P(x) > 0\) for all \(x\), while \(P_\sigma(x)\) (\(\sigma \neq \text{Id}\)) takes both positive and negative values. Therefore, such positive reals \(a_0, a_1, \ldots, a_{19}\) do exist.

The answer is: \boxed{\text{Yes}}.
\end{tcolorbox}}
    \caption{Sample problem of the subset of algebra questions of Omni-Math.}
    \label{fig:sample-3}
\end{figure*}

\begin{figure}
    \centering
    \includegraphics{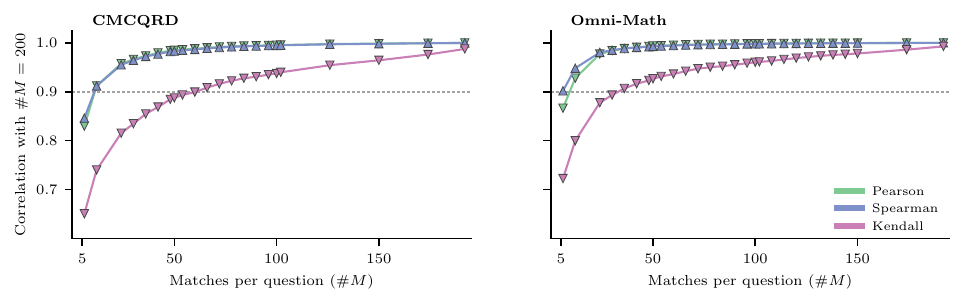}
    \caption{OpenAI o3 and Gemini 2.5 Pro play $66$ matches per problem for CMCQRD and $36$ matches per problem for Omni-Math (see \Cref{subsec:Compare}). For large datasets, it is impossible to compare every pair of problems. Therefore, we use the smaller gpt-oss-120b model to determine when the Bradley–Terry scores converge as a function of the number of matches played per problem. Due to money and time constraints, we take $200$ as the maximum possible number of matches. Then, we compute the correlation between $\#M=200$ and $\#M=6,12,24,30,36,...,192$. Pearson and Spearman correlations converge quickly, while Kendall's tau is stricter, as we have previously observed. We set the threshold at Kendall $\tau > 0.90$ and $>0.95$ for the other coefficients. For JEE ($n=34$), we play all the matches. }
    \label{fig:FigureA4}
\end{figure}

\clearpage

\begin{figure}
    \centering
    \includegraphics{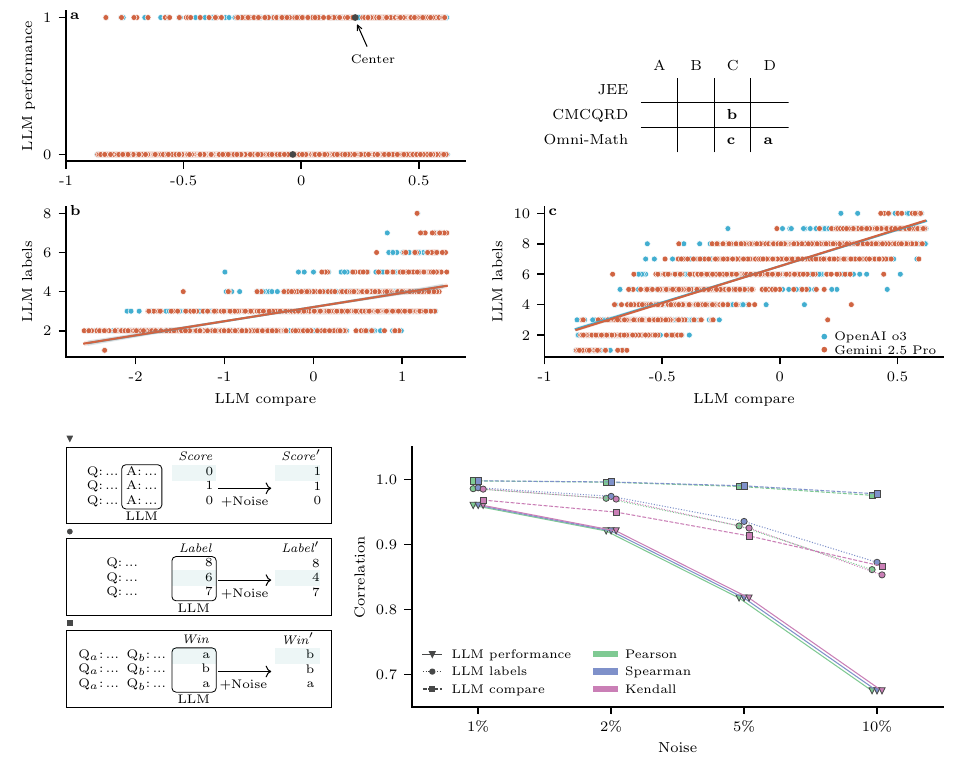}
    \caption{The main results on LLM-based measures (see \Cref{subsec:correlation}) continue to hold when we change the employed language model. In this figure, we use \emph{Gemini 2.5 Pro} instead of OpenAI o3 to construct the LLM-based measures (\Cref{fig:Figure2}) and to test their robustness (\Cref{fig:Figure3}). (Top) LLM compare (our method) correlates positively with LLM labels and LLM performance across datasets. (Bottom) Of all LLM-based measures, LLM compare is the most robust to hallucinations.   }
    \label{fig:FigureA6}
\end{figure}

\end{document}